\def\year{2019}\relax
\documentclass[letterpaper]{article} 
\usepackage{aaai19}  
\usepackage{times}  
\usepackage{helvet}  
\usepackage{courier}  
\usepackage{url}  
\usepackage{graphicx}  
\usepackage{booktabs}       
\usepackage{comment}

\usepackage{rotating}
\usepackage{multirow}
\usepackage{varwidth}
\usepackage[lofdepth,lotdepth]{subfig}

\usepackage{amsmath,amssymb} 
\usepackage{bm}
\usepackage{amssymb}
\usepackage{pifont}
\usepackage{lipsum}
\usepackage{xcolor}
\usepackage{array}

\newcommand{\cmark}{\ding{51}}%
\newcommand{\xmark}{\ding{55}}%

\DeclareMathOperator*{\Rank}{Rank}

\newcommand{\tens}[1]{
\bm{\mathcal{#1}}
}

\begin{document}
%
\title{BLOCK: Bilinear Superdiagonal Fusion for Visual Question Answering and Visual Relationship Detection}

\author{Hedi Ben-younes$^{1,2}$ \ Remi Cadene$^1$ \ Nicolas Thome$^3$ \ Matthieu Cord$^1$ \\
$^1$Sorbonne Universit\'e, CNRS, LIP6, F-75005 Paris, France \\ $^2$Heuritech \\
$^3$CEDRIC - Conservatoire National des Arts et M\'etiers, 75003 Paris,
France
}

\maketitle
\begin{abstract}
 Multimodal representation learning is gaining more and more interest within the deep learning community. While bilinear models provide an interesting framework to find subtle combination of modalities, their number of parameters grows quadratically with the input dimensions, making their practical implementation within classical deep learning pipelines challenging. 
 In this paper, we introduce BLOCK, a new multimodal fusion based on the block-superdiagonal tensor decomposition. It leverages the notion of block-term ranks, which generalizes both concepts of rank and mode ranks for tensors, already used for multimodal fusion. 
 It allows to define new ways for optimizing the tradeoff between the expressiveness and complexity of the fusion model, and is able to represent very fine interactions between modalities while maintaining powerful mono-modal representations. 
 We demonstrate the practical interest of our fusion model by using BLOCK for two challenging tasks: Visual Question Answering (VQA) and Visual Relationship Detection (VRD), where we design end-to-end learnable architectures for representing relevant interactions between modalities.
 Through extensive experiments, we show that BLOCK compares favorably with respect to state-of-the-art multimodal fusion models for both VQA and VRD tasks. Our code is available at \url{https://github.com/Cadene/block.bootstrap.pytorch}.
 \end{abstract}

\section{Introduction}
In many of the most recent tasks tackled by artificial intelligence, multiple sources of information need to be taken into account for decision making. Problems such as visual question answering \cite{Goyal_2017_CVPR}, visual relationship detection \cite{VRD_Lu_2016_ECCV}, cross-modal retrieval \cite{kiros2015unifying} or social-media post classification \cite{Duong2017MultimodalCF} require, at a certain level and to some extent, the fusion between multiple modalities.

For classical mono-modal tasks, linear models constitute a handful building block to transform the input $\bm{x}$ and to match it with the desired output $\bm{y}$. When dealing with two modalities, not only do we need to properly transform each input $\bm{x}_1$ and $\bm{x}_2$ into a representation that fits the problem, we also want to model the interactions between these modalities. 
A natural candidate to extend linear models for two inputs are bilinear models. However, 
the number of parameters of a bilinear model is quadratic in the input dimensions: as a linear model is characterized by a matrix $\bm{W} \in \mathbb{R}^{dim(\bm{x}) \times dim(\bm{y})}$, a bilinear model is defined by a tensor $\tens{T} \in \mathbb{R}^{dim(\bm{x}_1) \times dim(\bm{x}_2) \times dim(\bm{y})}$. When the input dimensions grow  (thousand or more), learning a full tensor $\tens{T}$ becomes quickly intractable.
The main issue is to reduce and control the numbers of parameters representing $\tens{T}$. 
Usually, when working with linear models, restricting the complexity is done by constraining the rank of the matrix $\bm{W}$. Unfortunately, it is much more complicated when it comes to bilinear models, since it requires notions of multi-linear algebra. While a lot of work has been done in extending the notion of complexity to higher-order tensors \cite{Harshman2001FoundationsOT}, \cite{Carroll1970}, \cite{Tucker1966}, it is not clear whether the simple extension of \textit{rank} should be used to constrain a bilinear model. 

In this paper, we tackle the general problem of learning end-to-end bilinear models. We propose BLOCK, a Block Superdiagonal Fusion framework for multimodal representation based on the block-term tensor decomposition \cite{De08f}. As it has been studied in the signal processing literature~\cite{Cichocki2015TensorDF}, the focus was on having uniqueness properties to ensure a physically interpretable decomposition. We study here this decomposition under a machine learning perspective instead, and use it as a fully learnable tensor of parameters. Interestingly, this decomposition leverages the notion of block-term ranks to define a tensor's complexity. It encapsulates both concepts of \textit{rank} and \textit{mode ranks}, at the basis of Candecomp/PARAFAC \cite{Harshman2001FoundationsOT} and Tucker decomposition \cite{Tucker1966}. This complexity analysis is capitalized on to provides a new way to control the tradeoff between the expressiveness and complexity of the fusion model. More precisely, BLOCK enables to model very rich (\textit{i.e.} full bilinear) interactions between groups of features, while the block structure limits the whole complexity of the model, which enables to keep expressive (\textit{i.e.} high dimensional) mono-moodal representations. 

The BLOCK model is used for solving  two challenging applications: Visual Question Answering (VQA) and Visual Relationship Detection (VRD). 
For both tasks, the number of blocks and the size of each projection in the BLOCK fusion will be adapted to balance between fine interaction modeling and low number of parameters.
We embed our bilinear BLOCK fusion strategy into deep learning architectures ; through extensive experiments, we validate the relevance of the approach as we provide an extensive and systemic comparison of many state-of-the-art multimodal fusion techniques. Moreover, we obtain very competitive results 
on three commonly used datasets: VQA 2.0 \cite{Goyal_2017_CVPR}, TDIUC \cite{Kafle_2017_ICCV} and the VRD dataset \cite{VRD_Lu_2016_ECCV}.

\section{BLOCK fusion model}
\label{section:block}
In this section, we present our BLOCK fusion strategy and discuss its connection to other bilinear fusion methods from the literature.

A bilinear model takes as input two vectors $\bm{x}^1 \in \mathbb{R}^{I}$ and $\bm{x}^2 \in \mathbb{R}^{J}$, and projects them to a K-dimensional space with tensor products:
\begin{equation}
 \label{eq:bilinear}
 \bm{y} = \tens{T} \times_1 \bm{x}^1 \times_2 \bm{x}^2
\end{equation}
where $\bm{y} \in \mathbb{R}^K$. Each component of $\bm{y}$ is a quadratic form of the inputs: $\forall k \in [1, K]$,
\begin{equation}
 y_k = \sum_{i=1}^I \sum_{j=1}^J \tens{T}_{ijk} . x^1_i . x^2_j
\end{equation}
A bilinear model is completely defined by its associated tensor $\tens{T} \in \mathbb{R}^{I \times J \times K}$, the same way as a linear model is defined by its associated matrix. 

\subsection{BLOCK model}
\label{subsection:block}
\begin{figure}
 \includegraphics[width=\linewidth]{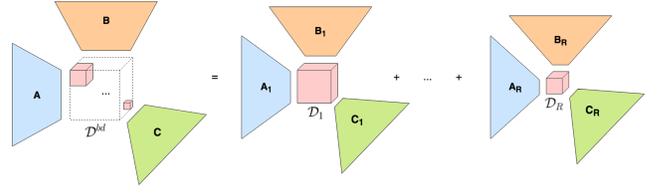}
 \caption{\label{fig:blocktucker}
 BLOCK framework. The third-order interaction tensor is decomposed in $R$ rank-(L,M,N) terms. We give here the two equivalent representations of the block-term decomposition, presented in this paper. On the left, we show the formulation with the block-superdiagonal tensor decomposition that corresponds to Eq.~(\ref{eq:bd}). On the right, we express it as a sum of small decompositions, as written in Eq.~(\ref{eq:sum}). Through the $R$ number of blocks and the dimensions of the $\bm{A}, \bm{B}$ and $\bm{C}$ projections, we can handle the trade-off between model complexity and expressivity.}
\end{figure}

In order to reduce the number of parameters and constrain the model's complexity, we express $\tens{T}$ using the block-term decomposition. 
More precisely, the decomposition of $\tens{T}$ in rank (L,M,N) terms is defined as:
\begin{equation}
 \label{eq:sum}
 \tens{T} := \sum_{r=1}^R \tens{D}_r \times_1 \bm{A}_r \times_2 \bm{B}_r \times_3 \bm{C}_r
\end{equation}
where $\forall r\in [1, R]$, $\tens{D}_r \in \mathbb{R}^{L\times M \times N}$, $\bm{A}_r \in \mathbb{R}^{I \times L}, \bm{B}_r \in \mathbb{R}^{J \times M}$ and $\bm{C}_r \in \mathbb{R}^{K \times N}$.
This decomposition is called \textit{block-term} because it can be written as  
\begin{equation}
 \label{eq:bd}
 \tens{T} = \tens{D}^{bd} \times_1 \bm{A} \times_2 \bm{B} \times_3 \bm{C}
\end{equation}
where $\bm{A} = [\bm{A}_1, ..., \bm{A}_R]$ (same for $\bm{B}$ and $\bm{C}$), and $\tens{D}^{bd} \in \mathbb{R}^{LR \times MR \times NR}$ the block-superdiagonal tensor of $\{\tens{D}_r\}_{1\leq r \leq R}$, as illustrated in Figure~\ref{fig:blocktucker}.
Applying this structural constraint to $\tens{T}$ in Eq.~(\ref{eq:bilinear}), we can express $\bm{y}$ with respect to $\bm{x}^1$ and $\bm{x}^2$. Let $ \hat{\bm{x}}^1 = \bm{A} \bm{x}^1 \in \mathbb{R}^{LR}$ and $\hat{\bm{x}}^2 = \bm{B} \bm{x}^2 \in \mathbb{R}^{MR}$. These two projections are merged with a fusion parametrized by the block-superdiagonal tensor $\tens{D}^{bd}$. Each block in this tensor merges together chunks of size $L$ from $\hat{\bm{x}}^1$ and of size $M$ from $\hat{\bm{x}}^2$ to produce a vector of size $N$:
\begin{equation}
 \bm{z}_r = \tens{D}_r \times_1 \hat{\bm{x}}^1_{rL:(r+1)L} \times_2 \hat{\bm{x}}^2_{rM:(r+1)M}
\end{equation}
where $\hat{\bm{x}}_{i:j}$ is a vector of dimension $j-i$ containing the corresponding values in $\hat{\bm{x}}$.
Finally, all the $\bm{z}_r$ vectors are concatenated to produce $\bm{z} \in \mathbb{R}^{NR}$. The final prediction vector is $\bm{y} = \bm{C} \bm{z} \in \mathbb{R}^{K}$. 

To further reduce the number of parameters in the model, we add a constraint on the rank of each third order slices matrices of the blocks $\tens{D}_r$, as it was done in some recent VQA applications (see Section~\ref{section:vqa_exp}).

\paragraph{Discussion}
When working with a linear model, a usual technique to restrict the hypothesis space and number of parameters is to constrain the rank of its associated matrix. The formal notion of \textit{matrix rank} quantifies how complex a linear model is allowed to be. However, when it comes to restricting the complexity of a bilinear model, multiple algebraic concepts can be used. We give two examples that are related to the block-term decomposition.

The Candecomp/PARAFAC (CP) decomposition \cite{Carroll1970}, \cite{Harshman2001FoundationsOT} of a tensor $\tens{T} \in \mathbb{R}^{I \times J \times K}$ is the linear combination of rank-1 terms 
\begin{equation}
 \label{eq:cp}
 \tens{T} := \sum_{r=1}^R \bm{a}_r \circ \bm{b}_r \circ \bm{c}_r
\end{equation}
where $\circ$ denotes the outer product, and the vectors $\bm{a}_r \in \mathbb{R}^I$, $\bm{b}_r \in \mathbb{R}^J$ and $\bm{c}_r \in \mathbb{R}^K$ represent the elements of the decomposition. The rank of $\tens{T}$ is defined by the minimal number $R$ of triplet vectors so that the Eq.~(\ref{eq:cp}) is true.
Thus, restricting the hypothesis space for $\tens{T}$ to the set of tensors defined by Eq.~(\ref{eq:cp}) guarantees that the rank is upper-bounded by $R$. 

Applying this constraint on $\tens{T}$, Eq.~(\ref{eq:bilinear}) is simplified into
\begin{equation}
 \bm{y} = \bm{C}\left( \left( \bm{x}^{1 \top } \bm{A} \right) \ast \left( \bm{x}^{2 \top} \bm{B} \right) \right)
\end{equation}
where $\bm{A} = [\bm{a}_1, ..., \bm{a}_R] \in \mathbb{R}^{I \times R}$, $\bm{B} = [\bm{b}_1, ..., \bm{b}_R] \in \mathbb{R}^{J \times R}$ and $\bm{C} = [\bm{c}_1, ..., \bm{c}_R] \in \mathbb{R}^{K \times R}$, and $\ast$ denotes element-wise product. This decomposition
 can be seen as a special case of the block-term decomposition where $L=M=N=1$, reducing $\tens{D}^{bd}$ to a super-diagonal identity tensor.

Another way to restrict a three-way tensor's complexity is through its \textit{mode ranks}. The rank-(L,M,N) Tucker decomposition \cite{Tucker1966} of $\tens{T} \in \mathbb{R}^{I \times J \times K}$ is defined as
\begin{equation}
 \tens{T} := \tens{D} \times_1 \bm{A} \times_2 \bm{B} \times_3 \bm{C}
\end{equation}
where $\tens{D} \in \mathbb{R}^{L \times M \times N}$, $\bm{A} \in \mathbb{R}^{I \times L}$, $\bm{B} \in \mathbb{R}^{J \times M}$ and $\bm{C} \in \mathbb{R}^{K \times N}$. This decomposition assumes a constraint on the three unfolding matrices of $\tens{T}$, such that $\Rank \left(\tens{T}_{JK \times I}\right)~=~L$, $\Rank \left(\tens{T}_{KI \times J}\right) = M$ and $\Rank \left(\tens{T}_{IJ \times K}\right) = N$ (following the notations in \cite{De08f}).

Applying this constraint to $\tens{T}$, Eq.~(\ref{eq:bilinear}) can be re-written as:
\begin{equation}
 \bm{y} = \bm{C} \left( \tens{D} \times_1 \left( \bm{x}^{1 \top} \bm{A} \right) \times_2 \left( \bm{x}^{2 \top} \bm{B} \right) \right)
\end{equation}
This decomposition can be seen as a special case of the block-term decomposition where there is only $R=1$ block in the core tensor.
    
As was studied by \cite{De08f}, the notion of tensor complexity should be expressed not only in terms of rank or mode ranks, but using the number of blocks and the mode-n ranks of each block. It appears that CP and Tucker decompositions are two extreme cases, where only one of the two quantities is used. For the CP, the number of blocks corresponds to the rank, but each block is of size~(1,1,1). 
The monomodal projections can be high-dimensional and thus integrate rich transformation of the input, but the interactions between both projections is relatively poor as a dimension from one space is only allowed to interact with another. For the Tucker decomposition, there is only one block of size (L,M,N). The interaction modeling is very rich since all inter-correlations between feature dimensions of the different modalities are considered. However, this quantity of possible interactions limits the dimensions of the projected space, which can cause a bottleneck in the model.

BLOCK being built on the block-term decomposition, we constrain the tensor using a combination of both concepts, which provides a richer modeling of the interactions between modalities. This richness is ensured by the $R$ tensors $\tens{D}_r$, each parametrizing a bilinear function that takes as inputs chunks of $\tilde{\bm{x}}^1$ and $\tilde{\bm{x}}^2$. As this interaction modelling is done by chunks and not for every possible combination of components in $\tilde{\bm{x}}^1$ and $\tilde{\bm{x}}^2$, we can reach high dimensions in the projections $\bm{A}$ and $\bm{B}$ without exploding the number of parameters in $\tens{D}^{bd}$. 
This property of having a fine interaction modeling between high dimensional projections is very desirable in our context where we need to model complex interactions between high-level semantic spaces. As we show in the experiments, performance of a bilinear model strongly depends on both the number and the size of the blocks $\tens{D}_r$ that parametrize the system.

\section{BLOCK fusion for VQA task}
\label{section:vqa_exp}

The task of Visual Question Answering \cite{VQA}, \cite{Goyal_2017_CVPR} has been a fertile playground for researchers to investigate on how bilinear models should be used. 
In the classical setup for VQA, shown in Figure~\ref{fig:vqa_fusion}, image and question have to be merged with a multimodal fusion technique, which can be implemented as an instance of bilinear model. The answer is predicted through a classification layer that follows the fusion module. 
In MCB \cite{FukuiPYRDR16}, the bilinear interaction is simplified using a sketching technique. However, more recent techniques tackle this complexity issue from a tensor decompositions standpoint: MLB \cite{Kim2017} and MUTAN \cite{benyounescadene2017mutan} constrain the tensor of parameters using respectively the CP and Tucker decomposition. In MFB \cite{yu2017mfb}, the tensor is viewed as a stack of matrices, and a classical matrix rank constraint is imposed on each of them. Finally in MFH \cite{yu2018beyond}, multiple MFB blocks are cascaded to model higher-order interactions between inputs.
Most recent techniques embed some of these fusion strategies into more elaborated architecture that involve multiple types of visual features \cite{pythia2018}, or specific modules \cite{zhang2018learning} designed for precise question types.
In this section, we compare BLOCK to the other multimodal fusion techniques, and show how it surpasses them both in terms of performance and number of parameters. 

\subsection{VQA architecture}

\begin{figure}
    \centering
    \includegraphics[width=\linewidth]{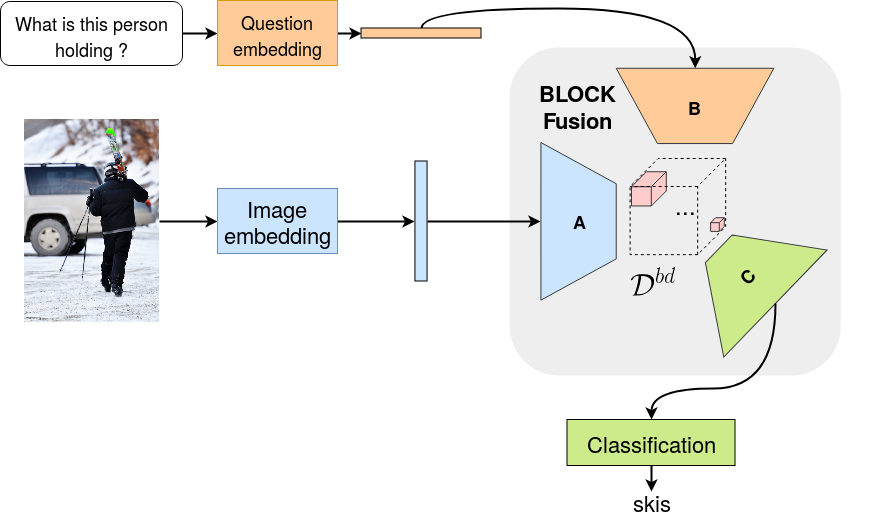}
    \caption{\label{fig:vqa_fusion} Architecture for VQA that embeds the BLOCK bilinear fusion. To make this system more efficient, we integrate the fusion in an attentional framework.}
\end{figure}

Our VQA model is based on a classical attentional architecture \cite{FukuiPYRDR16}, enriched by our proposed merging scheme. Our fusion model is shown in Figure~\ref{fig:vqa_fusion}. 
We use the Bottom-up image features provided by \cite{Teney_2018_CVPR}, consisting of a set of detected objects and their representation (see \cite{bmvc17,wildcat} for further insights on detection and localization). 
To get a vector embedding of the question, words are preprocessed and then fed into a pretrained Skip-thought encoder \cite{Kiros_2015_NIPS}.
The outputs of this language model are used to produce a single vector representing the whole question, as in \cite{yu2018beyond}.
We use a BLOCK fusion to merge the question and image representations.
The question vector is used as a context to guide the visual attention. Saliency scores are produced using a BLOCK fusion between each image vector and the question embedding.

~

\textbf{Details}: For the BLOCK layers, we set $L = M = N = 80, R = 20$ and constrain the rank of each mode-3 slices of each block to be less than 10. We found these hyperparameters with a a cross-validation on the \textit{val} set.
As in \cite{yu2018beyond}, we consider the 3000 most frequent answers. As in \cite{benyounescadene2017mutan}, we use a cross-entropy loss with answer sampling.
We jointly optimize the parameters of our VQA model using Adam \cite{KingmaB14} with a learning rate of $1e^{-4}$, without learning rate decay or gradient clipping, and with a batch size of 200. We early stop the training of our models according to their accuracy on a holdout set.

\subsection{Fusion analysis}
In Table~\ref{tab:vqa2_fusion}, we compare BLOCK to 8 different fusion schemes available in the literature on the commonly used VQA 2.0 Dataset \cite{Goyal_2017_CVPR}. This dataset is composed of 658,111 image-question-answers triplets for training and validation (\textit{trainval} set), and 447,793 triplets for evaluation (\textit{test-std} set). We train on \textit{trainval} minus a small subset used for early-stopping, and report the performance on \textit{test-dev} set. For each fusion strategy, we run a grid search over its hyperparameters and keep the model that performs best on our validation set. We report the size of the model, corresponding to the number of parameters between the attended image features, the question embedding, and the answer prediction. We briefly describe the different fusion schemes used for the comparison:

-- (1) the two vectors are projected on a common space, and their summation is projected to predict the answer; 

-- (2) the vectors are concatenated and passed at the input of a 3-layer MLP; 

-- (3) a bilinear interaction based on a count-sketching technique that projects the outer product of between inputs on a multimodal space;

-- (4) a bilinear interaction where the tensor is expressed as a Tucker decomposition;

-- (5) a bilinear interaction where the tensor is expressed as a CP decomposition;

-- (6) a bilinear interaction where each 3rd mode slice matrix of the tensor is constrained by its rank;

-- (7) a bilinear interaction where the tensor is expressed as a Tucker decomposition, and where its core tensor has the same rank constraint as (6);

-- (8) a higher order fusion composed of cascaded (6);

-- (9) our BLOCK fusion.

From the results in Table~\ref{tab:vqa2_fusion}, we see that the simple sum fusion (1) provides a very low baseline. We also note that the MLP (2) doesn't provide the best results, despite its non-linear structure. As the MLP should be able to find that two different modalities are used and that it needs to look for interactions between them, this is in practice difficult to obtain. Instead, top performing methods are based on a bilinear model. The structure imposed on the parameters highly influences the final performance. We can see that (3), which simplifies the bilinear model using random projections, has efficiency issues due to the count-sketching technique. These issues are alleviated in the other bilinear methods, which use the tensor decomposition framework to practically implement the interaction. Our BLOCK method (9) gives the best results. As we saw, the block-term decomposition generalizes both CP and Tucker decompositions, which is why it is not surprising to see it surpass them. Moreover, the fact that it integrates the 3rd order slices rank constraint gives it the advantages of (6) and (7). Interestingly, it even surpasses (8) which is based on a higher-order interaction modeling, while using 30M less parameters. This strongly indicates that controlling a bilinear model through its block-term ranks provides an efficient trade-off between modeling capacities and number of parameters. To further validate this hypothesis, we evaluate a BLOCK fusion with only 3M parameters. This model obtained 64.91\%. Unsurprisingly, it does not surpasses all the methods against which we compare. However, it obtains competitive results, improving over 5 out of 8 methods that all use far more parameters. 

\newcolumntype{L}[1]{>{\raggedright\let\newline\\\arraybackslash\hspace{0pt}}m{#1}}
\newcolumntype{C}[1]{>{\centering\let\newline\\\arraybackslash\hspace{0pt}}m{#1}}
\begin{table*}[t]
    \centering
    \caption{\label{tab:vqa2_fusion} Comparison of the fusion schemes on VQA2 \textit{test-dev} set. $|\Theta|$ is the number of parameters learned in the fusion modeling. \textit{All} is the overall Open Ended accuracy (higher is better). \textit{Yes/no}, \textit{Numbers} and \textit{Others} are subsets that correspond to answers types. In the descriptions, the letter B corresponds to a bilinear model. }
    \begin{tabular}{l c *6{c}}
        \toprule
	    & Description & Reference & $|\Theta|$ & All & Yes/no & Number & Other \\
        \midrule
        (1) & Linear & Sum & 8M & 58.48 & 71.89 & 36.56 & 52.09 \\
        (2) & Non-linear & Concat MLP & 13M & 63.85 & 81.34 & 43.75 & 53.48 \\
        (3) & B + count-sketching & MCB \cite{FukuiPYRDR16} & 32M & 61.23 & 79.73 & 39.13 & 50.45 \\
        (4) & B + Tucker decomp. & Tucker \cite{benyounescadene2017mutan} & 14M & 64.21 & 81.81 & 42.28 & 54.17 \\
        (5) & B + CP decomp. & MLB \cite{Kim2017} & 16M & 64.88 & 81.34 & 43.75 & 53.48 \\
        (6) & B + low-rank on the 3rd mode slices & MFB \cite{Yu_2017_ICCV} & 24M & 65.56 & 82.35 & 41.54 & 56.74 \\
        (7) & Combination of (4) and (6) & MUTAN \cite{benyounescadene2017mutan} & 14M & 65.19 & 82.22 & 42.1 & 55.94 \\
        (8) & Higher order fusion & MFH \cite{yu2018beyond} & 48M & 65.72 & 82.82 & 40.39 & 56.94 \\
        \midrule
        (9) & B + Block-term decomposition & BLOCK & 18M & \textbf{66.41} & \textbf{82.86} & \textbf{44.76} & \textbf{57.3} \\ 
        \bottomrule
    \end{tabular}
\end{table*}

\subsection{Comparison to leading VQA methods}

We compare our model with state-of-the-art VQA architecture on two datasets: the widely used VQA 2.0 Dataset \cite{Goyal_2017_CVPR} and TDIUC \cite{Kafle_2017_ICCV}. On this more recent dataset, evaluation metrics are provided to assess the robustness of the model with respect to answer imbalance, as well as to account for performance homogeneity across the difference question types. 

\begin{table*}[t]
    \centering
    \caption{\label{tab:tdiuc} State-of-the-art comparison on the TDIUC testing set. * scores reported from \cite{Kafle_2017_ICCV}.}
    \begin{tabular}{c c c c c c}
        \toprule
        Model & Accuracy & A-MPT & H-MPT & A-NMPT & H-NMPT \\
        \midrule
        Most common answer \cite{Kafle_2017_ICCV} & 51.15 & 31.11 & 17.53 & 15.63 & 0.83 \\
        Question only \cite{Kafle_2017_ICCV} & 62.74 & 39.31 & 25.93 & 21.46 & 8.42 \\
        NMN* \cite{Andreas_2016_CVPR} & 79.56 & 62.59 & 51.87 & 34.00 & 16.67 \\
        MCB* \cite{FukuiPYRDR16} & 81.86 & 67.90 & 60.47 & 42.24 & 27.28 \\
        RAU* \cite{Noh_2016_Arxiv} & 84.26 & 67.81 & 59.00 & 41.04 & 23.99 \\
        \midrule
        BLOCK & \textbf{85.96} & \textbf{71.84} & \textbf{65.52} & \textbf{58.36} & \textbf{39.44} \\
        \bottomrule
    \end{tabular}
\end{table*}

As we show in Table~\ref{tab:tdiuc}, our model is able to outperform the preceding ones on TDIUC by a large margin for every metrics, especially those which account for bias in the data. We notably report a gain of +1.7 in accuracy, +3.95 in A-MPT, +5.05 in H-MPT, +16.12 in A-NMPT, +15.45 in H-NMPT, over the best scoring model in each metric. The high results in the harmonic metrics (H-MPT and H-NMPT) suggest that BLOCK performs well across all question types, while the high scores in the normalized metrics (A-NMPT and H-NMPT) denote that our model is robust to answer imbalance type of bias in the dataset.

In Table~\ref{tab:vqa2}, we see that our fusion model obtains competitive results on VQA 2.0 compared to previously published methods. As we are outperformed by \cite{zhang2018learning}, whose proposition rely on a completely different architecture, we believe that both our contributions are orthogonal. 
Still, our model performs better than \cite{Teney_2018_CVPR} and \cite{yu2018beyond}, with whom we share the global VQA architecture. 
In further details, we point out that BLOCK surpasses \cite{yu2018beyond} reaching a +1.78 improvement in the overall accuracy on \textit{test-dev}, even though the latter encompasses the current state-of-the-art fusion scheme. Furthermore, we use the same image features than \cite{Teney_2018_CVPR} and are able to achieve a +2.26 gain on \textit{test-dev} and +2.25 on \textit{test-std}.

\newcommand\CustomC[1]{%
  \multirow{3}*{%
    \begin{varwidth}{5em}
    \center #1%
    \end{varwidth}}}

\addtolength{\tabcolsep}{-2.5pt} 
\begin{table*}[t]
    \centering
    \caption{\label{tab:vqa2} State-of-the-art results on VQA2 testing sets. The models were trained on the union of VQA 2.0 \textit{trainval} split and VisualGenome \cite{Krishna_2017_IJCV} train split. \textit{All} is the overall OpenEnded accuracy (higher is better). \textit{Yes/no}, \textit{Numbers} and \textit{Others} are subsets that correspond to answers types. Only single model scores are reported. * scores reported from \cite{Goyal_2017_CVPR}
    }
    \begin{tabular}{c c@{\hskip 0.1in} cccc c@{\hskip 0.1in} cccc}
        \toprule
        \CustomC{Model} & & \multicolumn{4}{c}{VQA2 Test-dev} && \multicolumn{4}{c}{VQA2 Test-std} \\
        \cmidrule{3-6} \cmidrule{8-11}
	    && All & Yes/no & Num. & Other && All & Yes/no & Num. & Other \\
        \midrule
        Most common answer \cite{Goyal_2017_CVPR} && - & - & - & - && 25.98 & 61.20 & 0.36 & 1.17 \\
        Question only \cite{Goyal_2017_CVPR} && - & - & - & - && 44.26 & 67.01 & 31.55 &  27.37 \\
        Deep LSTM* \cite{lu2015deeper} && - & - & - & - && 54.22 & 73.46 & 35.18 & 41.83  \\
        MCB* \cite{FukuiPYRDR16} && - & - & - & - && 62.27 & 78.82 & 38.28 & 53.36 \\
        ReasonNet \cite{Ilievski_2017_NIPS} &&  - & - & - & - && 64.61 & 78.86 & 41.98 & 57.39 \\
        TipsAndTricks \cite{Teney_2018_CVPR} && 65.32 & 81.82 & 44.21 & 56.05 && 65.67 & 82.20 & 43.90 & 56.26 \\
        MFH \cite{yu2018beyond} && 65.80 & - & - & - && - & - & - & - \\
        Counter \cite{zhang2018learning} && \textbf{68.09} & 83.14 & \textbf{51.62} & \textbf{58.97} && \textbf{68.41} & 83.56 & \textbf{51.39} & \textbf{59.11} \\
        \midrule 
        BLOCK && 67.58 & \textbf{83.6} & 47.33 & 58.51 && 67.92 & \textbf{83.98} & 46.77 &  58.79 \\
        \bottomrule
    \end{tabular}
\end{table*}
\addtolength{\tabcolsep}{2.5pt}

\section{VRD task}
\label{section:vrd_exp}
The task of Visual Relationship Detection aims at predicting triplets of the type \textit{"subject-predicate-object"} where \textit{subject} and \textit{object} are localized objects, and \textit{predicate} is a label corresponding to the relationship that links them (for example: "man-riding-bicycle", "woman-holding-phone"). To predict this relationship, multiple types of information are available, for both the subject and object regions: classes, bounding box coordinates, visual features, \textit{etc.} However, this context being more recent than VQA, fusion techniques are less formalized and more \textit{ad-hoc}. 
In \cite{Zhang_2017_CVPR}, the relation is predicted by a substractive fusion between subject and object representations, each consisting in a linear function of relative coordinates, class distributions and visual features. \cite{Li_2017_CVPR} predicts the relationship by a complex message passing structure between subject and object representations, and \cite{dai2017detecting} uses a formulation inspired from Conditional Random Fields to perform joint recognition between the subject, object and predicate classes.
We adopt in the following a very simple architecture, to put emphasis on the fusion module between different information sources. 

\subsection{VRD Architecture}
\begin{figure}
    \centering
    \includegraphics[width=\linewidth]{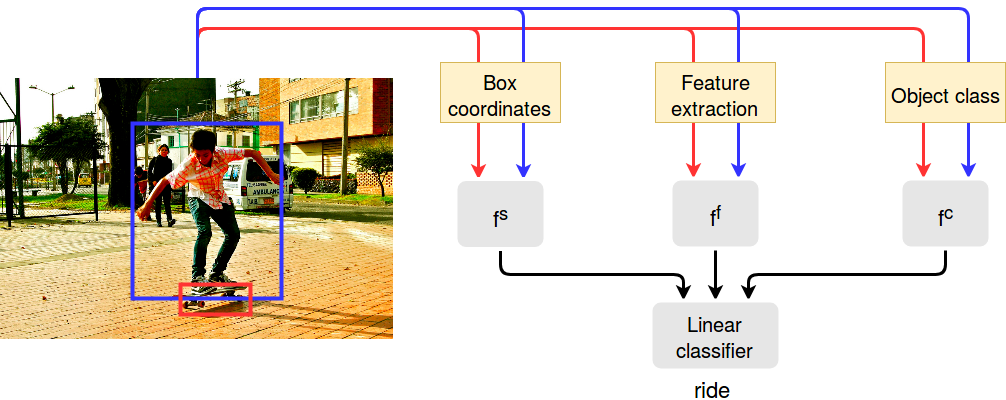}
    \caption{\label{fig:vrd_archi} Architecture of our visual relationship detection system}
\end{figure}

Our VRD architecture is shown in Figure~\ref{fig:vrd_archi}. It takes as inputs a subject and an object bounding box. Each of them is represented as their  4-dimensional box spatial coordinates $\bm{x}_s^{s}$ and $\bm{x}_o^{s}$ (normalized between 0 and 1), their object classes $\bm{x}_s^{c}$ and $\bm{x}_o^{c}$, and their semantic visual features $\bm{x}_s^{f}$ and $\bm{x}_o^{f}$. To predict the relationship predicate, we use one fusion module for each type of features following Eq.~(\ref{eq:vrd}).
\begin{equation}
 \label{eq:vrd}
 \bm{x} = [f^{s}\left(\bm{x}_s^{s}, \bm{x}_o^{s}\right),  f^{c}\left(\bm{x}_s^{c}, \bm{x}_o^{c}\right), f^{f}\left(\bm{x}_s^{f}, \bm{x}_o^{f}\right)]
\end{equation}
where $f$ can be implemented as BLOCK, or any other multimodal fusion. Each fusion module outputs a vector of dimension $d$, all concatenated into a 3d-dimensional vector that will serve as an input to a linear layer predictor $\bm{y} = \bm{W} \bm{x}$. The system is trained with back-propagation on a binary-crossentropy loss.

An other important component is the object detector. As usually done, we first train a Faster-RCNN on the object boxes of the VRD dataset. For Predicate prediction, we use it as a features extractor given the ground truth bounding boxes. For Phrase detection and Relationship detection, we use it to extract the bounding boxes with their associated features.

For a system to perform well on Phrase and Relationship detection, it should have been also trained on pairs of (subject, object) boxes that are not linked together by any relationship. During training, we randomly sample half of all possible \textit{negative pairs}, and assign them an all-zeros label vectors. 

\subsection{Dataset}
The VRD dataset \cite{VRD_Lu_2016_ECCV} is composed of 5,000 images with 100 object categories and 70 predicates. It contains 37,993 relationships with 6,672 unique triplets and an average of 24.25 predicates per object category. The dataset is divided between 4,000 images for training and 1,000 for testing.
Three different settings are commonly used to evaluate a model on VRD:
(1) \textbf{Predicate prediction}: the coordinates and class labels are given for both subject and object regions. This setup allows to assess the model's ability to predict a relationship, regardless of the object detection stage. 
(2) \textbf{Phrase detection}: a predicted triplet $<$\textit{subject, predicate, object}$>$ matches a ground-truth if the three labels match and if the union region of its bounding boxes matches the union region of the ground-truth triplet, with IoU above 0.5.
(3) \textbf{Relationship detection}: more challenging than (2), this one requires that both subject and object intersect with an IoU higher than 0.5. For each of these settings, performance is usually measured with Recall@50 and Recall@100

\subsection{Fusion analysis}
To show the effectiveness of the BLOCK bilinear fusion, we run the same type of experiment we did in the previous section. For each fusion technique, we use the architecture described in Equation~\ref{eq:vrd} where we replace $f$ by the corresponding bilinear function. As we did for VQA, we cross-validate the hyperparameters of each fusion technique and keep the best model each time.
In Table~\ref{tab:vrd_fusion}, we see that BLOCK still outperforms all previous methods on each of the three tasks. 
We can remark that for this task, the non linear MLP perform relatively well compared to the other methods. It is likely that an MLP can model the interactions at stake for VRD more easily than those for VQA. However, we can improve over this strong baseline using a BLOCK fusion. 

\newcommand\CustomB[1]{%
  \multirow{3}*{%
    \begin{varwidth}{8em}
    \center #1%
    \end{varwidth}}}

\begin{table*}[t]
    \centering
    \caption{\label{tab:vrd_fusion} Comparative study of the different multimodal fusion strategies on the VRD test-set. The reported metrics are the Recall@K in \%.}
    \begin{tabular}{c c c *3{c c}}
        \toprule
         &  \CustomB{Description} & \CustomB{$|\Theta|$} & \multicolumn{2}{c}{Predicate}& \multicolumn{2}{c}{Phrase}& \multicolumn{2}{c}{Relationship}\\
        & & &  \multicolumn{2}{c}{Prediction} &  \multicolumn{2}{c}{Detection} &  \multicolumn{2}{c}{Detection} \\
        \cmidrule(lr){4-9}
        & &  &                                     R@50 & R@100 
         &                                     R@50 & R@100
          &                                     R@50 & R@100\\
        \midrule
        (1) & Linear & 2.5M & 82.99 & 89.68 & 14.44 & 16.94 & 9.73 & 11.34 \\
        (2) & Non-linear & 2M & 84.47 & 91.6 & 21.9 & 24.69 & 15.79 & 17.83 \\
        (3) & B + count-sketching & 2M & 82.23 & 89.07 & 13.42 & 15.8 & 9.17 & 10.79\\
        (4) & B + Tucker decomp. & 3M & 83.25 & 89.77 & 11.23 & 14.09 & 7.37 & 9.00 \\
        (5) &  B + CP decomp. & 4M & 85.96 & 91.66  & 23.67 & 26.50 & 16.41 & 18.59 \\
        (6) &  B + low-rank on the 3rd mode slices & 15M & 85.21 & 91.06  & 25.31 & 28.03 & 17.83 & 19.77 \\
        (7) & Combination of (4) and (6) & 30M & 85.65 & 91.33  & 25.77 & 28.65 & 18.53 & 20.38 \\
        (8) & Higher order fusion & 16M & 85.58 & 91.3  & 26.09 & 28.73 & 18.81 & 20.63 \\
        \midrule 
        (9) & Block-term decomposition & 5M & \textbf{86.58} &\textbf{ 92.58 } & \textbf{26.32} & \textbf{28.96}  & \textbf{19.06} & \textbf{20.96}\\
        \bottomrule
    \end{tabular}
\end{table*}

\begin{figure}[t]
\centering
 \subfloat[][]{
  \label{fig:vrd_mlb_to_tucker}
  \includegraphics[width=0.49\linewidth]{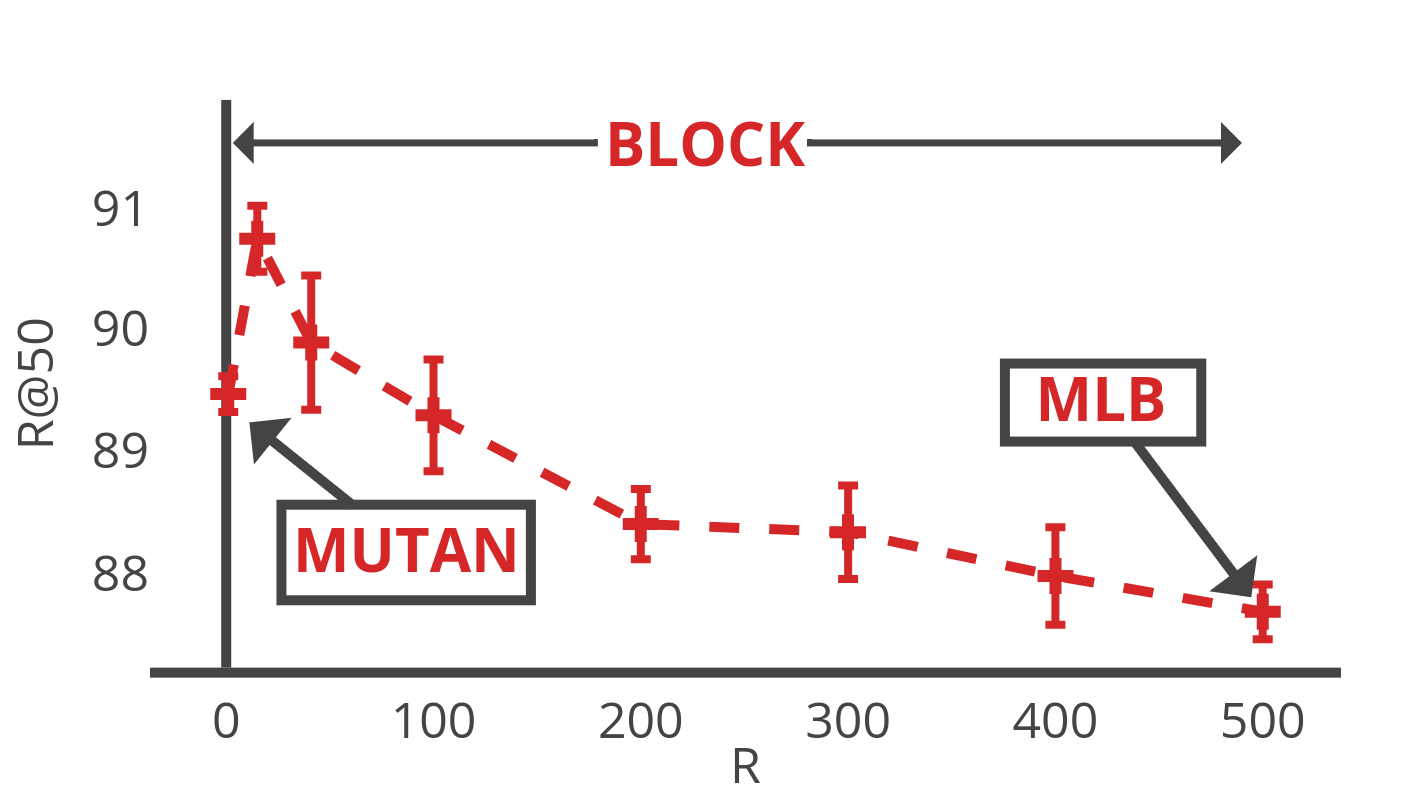}
 }
 \subfloat[][]{ 
  \label{fig:vrd_chunks}
  \includegraphics[width=0.49\linewidth]{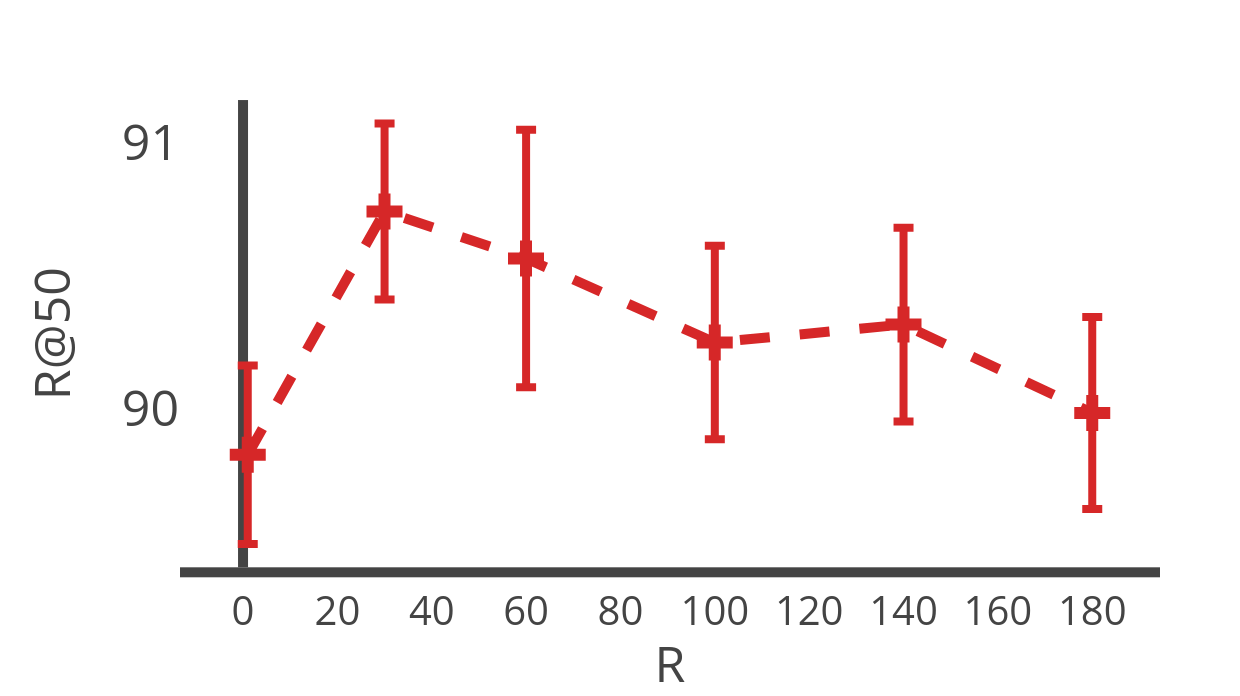}
}
  \caption{Performance of BLOCK with respect to the structure of the block-superdiagonal tensor. Scores are given on a holdout validation set. \ref{fig:vrd_mlb_to_tucker}: The size of the core tensor is fixed to $RL = RM = RN = 500$. \ref{fig:vrd_chunks}: The total number of parameters in the block-diagonal tensor is fixed to $555K$.}
\end{figure}

In the next experiments, we validate the power of our BLOCK fusion, and analyze how it behaves under different setups. We randomly split the training set into three train/val sets, and plot the mean and standard deviation of the recall@50 calculated over them. In Figure~\ref{fig:vrd_mlb_to_tucker}, we fix the dimension of the block-superdiagonal tensor to $RL = RM = RN = 500$ and vary the number of blocks used to fill this tensor. 
When $R=1$, which corresponds to the Tucker decomposition, the number of parameters in the core tensor is equal to $500^3 = 125M$, making the system arduously trainable on our dataset. On the opposite, when $R=500$, the number of parameters is controlled, but the mono-modal projections are only allowed to interact through an element-wise multiplication, which makes the interaction modeling relatively poor.
The block-term decomposition provides an in-between working regime, reaching an optimum when $R \approx 20$.

In Figure~\ref{fig:vrd_chunks}, we keep the number of parameters fixed. As the number of chunks increases, the dimensions of the mono-modal projections also increases. Once again, an optimum is reached when $R \approx 20$. These results confirm our hypothesis that the way the parameters are distributed within the tensor, in terms of size and number of blocks, has a real impact on the system's performance.

\subsection{Comparison to leading VRD methods}

\newcommand\CustomA[1]{%
  \multirow{3}*{%
    \begin{varwidth}{5em}
    \center #1%
    \end{varwidth}}}
\begin{table*}[t]
    \centering
    \caption{\label{tab:vrd} State-of-the-art results on the VRD testing set. The reported metrics are the Recall@K in \%.}
    \begin{tabular}{c c * 3{c c}}
        \toprule
        \CustomA{Model} & \CustomA{External \newline data} & \multicolumn{2}{c}{Predicate} & \multicolumn{2}{c}{Phrase} & \multicolumn{2}{c}{Relationship} \\
        &                                       & \multicolumn{2}{c}{Prediction} & \multicolumn{2}{c}{Detection} & \multicolumn{2}{c}{Detection} \\
        \cmidrule(lr){3-4} \cmidrule(lr){5-6} \cmidrule(lr){7-8} 
        &                                       & R@50 & R@100 & R@50 & R@100 & R@50 & R@100 \\
        \midrule
        Yu et. al \cite{Yu_2017_ICCV} & \cmark & 85.64 & 94.65 & 26.32 & 29.43 & 22.68 & 31.89 \\
        \midrule
        Li et. al \cite{Li_2017_CVPR}    & \xmark & - & - & 22.78 & 27.91 & 17.32 & 20.01 \\
        Liang et. al \cite{Liang_2017_CVPR} & \xmark & - & - & 21.37 & 22.60 & 18.19 & 20.79 \\
        Zhang et. al \cite{Zhang_2017_CVPR} & \xmark & 44.76 & 44.76 & 19.42 & 22.42 & 14.07 & 15.20 \\
        Lu et. al \cite{VRD_Lu_2016_ECCV} & \xmark & 47.87 & 47.87 & 16.17 & 17.03 & 13.86 & 14.70 \\
        Peyre et. al \cite{Peyre17} & \xmark & 52.6 & 52.6 & 17.9 & 19.5 & 15.8 & 17.1 \\
        Dai et. al \cite{dai2017detecting} & \xmark & 80.78 & 81.90 & 19.93 & 23.45 & 17.73 & 20.88 \\
        \midrule
        BLOCK & \xmark & \textbf{86.58} &\textbf{ 92.58 } & \textbf{26.32} & \textbf{28.96}  & \textbf{19.06} & \textbf{20.96} \\
        \bottomrule
    \end{tabular}
\end{table*}

In Table~\ref{tab:vrd}, we compare our system to the state-of-the-art methods on VRD. 
On predicate prediction, our fusion outperforms all previous methods on R@50, including \cite{Yu_2017_ICCV} that uses external data. On R@100, the BLOCK fusion is only marginally outperformed by \cite{Yu_2017_ICCV}, but we perform better than all methods that don't use extra data. These results validate the efficiency of the block-term decomposition to predict a predicate by fusing information coming from ground truth subject and object boxes.
On phrase detection, our BLOCK fusion achieves better results than all previous models in R@50. 
Notably, the scores obtained for phrase detection are lower than for predicate prediction, since the ground truth regions are not provided in this setup.
Finally, on relationship detection, BLOCK surpasses all previous methods without extra data in R@50, and gives similar performance than \cite{dai2017detecting} in R@100. The scores for relationship detection are lower than for phrase detection: in this setup, a prediction is positive if both subject and object boxes match the ground truth. On contrary, in phrase detection, the comparison between prediction and ground truth is done on the union between subject and object regions. 
Lastly, unlike some of the methods reported in Table~\ref{tab:vrd}, we do not fine-tune or adapt the detection network to the visual relationship tasks.

\section{Conclusion}
In this work, we introduce BLOCK, a bilinear fusion model whose tensor of parameters is structured using the block-term decomposition. BLOCK aims at optimizing the tradeoff between complexity and modeling capacities, and combines the strengths of the CP and Tucker decompositions. It offers the possibility to model rich interactions between groups of features, while still using high-dimensional mono-modal representations.

We apply BLOCK for two challenging computer vision tasks: VQA and VRD, where the parameters of our BLOCK fusion model are learned. Comparative experiments 
show that BLOCK improves over previous fusion schemes including linear, bilinear and non-linear models. We also show that BLOCK is able to maintain competitive performances with very compact parametrization. 

In future works, we plan to extend the BLOCK idea to other applications. In particular, we want to explore the use of multiple input and output modalities, and to apply BLOCK for interpreting and explaining the behaviour of the multimodal deep fusion model \cite{burger18,sigir18}.

\paragraph{Acknowledgements}
This work has been supported within the Labex SMART supported by French state funds managed by the ANR within the Investissements d’Avenir programme under reference ANR-11-LABX-65.

\bibliography{egbib}

\begin{thebibliography}{}

\bibitem[\protect\citeauthoryear{Andreas \bgroup et al\mbox.\egroup
  }{2016}]{Andreas_2016_CVPR}
Andreas, J.; Rohrbach, M.; Darrell, T.; and Klein, D.
\newblock 2016.
\newblock Neural module networks.
\newblock In {\em Proceedings of the IEEE Conference on Computer Vision and
  Pattern Recognition},  39--48.

\bibitem[\protect\citeauthoryear{Antol \bgroup et al\mbox.\egroup }{2015}]{VQA}
Antol, S.; Agrawal, A.; Lu, J.; Mitchell, M.; Batra, D.; Zitnick, C.~L.; and
  Parikh, D.
\newblock 2015.
\newblock {VQA}: {V}isual {Q}uestion {A}nswering.
\newblock In {\em ICCV}.

\bibitem[\protect\citeauthoryear{Ben-Younes \bgroup et al\mbox.\egroup
  }{2017}]{benyounescadene2017mutan}
Ben-Younes, H.; Cad{\`{e}}ne, R.; Thome, N.; and Cord, M.
\newblock 2017.
\newblock Mutan: Multimodal tucker fusion for visual question answering.
\newblock {\em ICCV}.

\bibitem[\protect\citeauthoryear{Carroll and Chang}{1970}]{Carroll1970}
Carroll, J.~D., and Chang, J.-J.
\newblock 1970.
\newblock Analysis of individual differences in multidimensional scaling via an
  n-way generalization of ``eckart-young'' decomposition.
\newblock {\em Psychometrika}.

\bibitem[\protect\citeauthoryear{Carvalho \bgroup et al\mbox.\egroup
  }{2018}]{sigir18}
Carvalho, M.; Cad{\`{e}}ne, R.; Picard, D.; Soulier, L.; Thome, N.; and Cord,
  M.
\newblock 2018.
\newblock Cross-modal retrieval in the cooking context: Learning semantic
  text-image embeddings.
\newblock In {\em SIGIR}.

\bibitem[\protect\citeauthoryear{Cichocki \bgroup et al\mbox.\egroup
  }{2015}]{Cichocki2015TensorDF}
Cichocki, A.; Mandic, D.~P.; Phan, A.~H.; Caiafa, C.~F.; Zhou, G.; Zhao, Q.;
  and Lathauwer, L.~D.
\newblock 2015.
\newblock Tensor decompositions for signal processing applications: From
  two-way to multiway component analysis.
\newblock {\em IEEE Signal Processing Magazine}.

\bibitem[\protect\citeauthoryear{Dai, Zhang, and Lin}{2017}]{dai2017detecting}
Dai, B.; Zhang, Y.; and Lin, D.
\newblock 2017.
\newblock Detecting visual relationships with deep relational networks.
\newblock In {\em CVPR}.

\bibitem[\protect\citeauthoryear{De~Lathauwer}{2008}]{De08f}
De~Lathauwer, L.
\newblock 2008.
\newblock Decompositions of a higher-order tensor in block terms --- part ii:
  Definitions and uniqueness.
\newblock {\em SIAM J. Matrix Anal. Appl.} 30(3):1033--1066.

\bibitem[\protect\citeauthoryear{Duong, Lebret, and
  Aberer}{2017}]{Duong2017MultimodalCF}
Duong, C.~T.; Lebret, R.; and Aberer, K.
\newblock 2017.
\newblock Multimodal classification for analysing social media.
\newblock {\em ECML-PKDD}.

\bibitem[\protect\citeauthoryear{Durand \bgroup et al\mbox.\egroup
  }{2017}]{wildcat}
Durand, T.; Mordan, T.; Thome, N.; and Cord, M.
\newblock 2017.
\newblock {WILDCAT:} weakly supervised learning of deep convnets for image
  classification, pointwise localization and segmentation.
\newblock In {\em CVPR}.

\bibitem[\protect\citeauthoryear{Engilberge \bgroup et al\mbox.\egroup
  }{2018}]{burger18}
Engilberge, M.; Chevallier, L.; P{\'{e}}rez, P.; and Cord, M.
\newblock 2018.
\newblock Finding beans in burgers: Deep semantic-visual embedding with
  localization.
\newblock In {\em CVPR}.

\bibitem[\protect\citeauthoryear{Fukui \bgroup et al\mbox.\egroup
  }{2016}]{FukuiPYRDR16}
Fukui, A.; Park, D.~H.; Yang, D.; Rohrbach, A.; Darrell, T.; and Rohrbach, M.
\newblock 2016.
\newblock In {\em EMNLP}.

\bibitem[\protect\citeauthoryear{Goyal \bgroup et al\mbox.\egroup
  }{2017}]{Goyal_2017_CVPR}
Goyal, Y.; Khot, T.; Summers-Stay, D.; Batra, D.; and Parikh, D.
\newblock 2017.
\newblock Making the v in vqa matter: Elevating the role of image understanding
  in visual question answering.
\newblock In {\em CVPR}.

\bibitem[\protect\citeauthoryear{Hanwang~Zhang}{2017}]{Zhang_2017_CVPR}
Hanwang~Zhang, Zawlin~Kyaw, S.-F. C. T.-S.~C.
\newblock 2017.
\newblock Visual translation embedding network for visual relation detection.
\newblock In {\em CVPR}.

\bibitem[\protect\citeauthoryear{Harshman \bgroup et al\mbox.\egroup
  }{2001}]{Harshman2001FoundationsOT}
Harshman, R.~A.; Ladefoged, P.; Reichenbach, H.; Jennrich, R.~I.; Terbeek, D.;
  Cooper, L.; Comrey, A.; Bentler, P.~M.; Yamane, J.; Vaughan, D.; and Jahnke,
  B.
\newblock 2001.
\newblock Foundations of the parafac procedure: Models and conditions for an
  "explanatory" multimodal factor analysis.

\bibitem[\protect\citeauthoryear{Ilievski and Feng}{2017}]{Ilievski_2017_NIPS}
Ilievski, I., and Feng, J.
\newblock 2017.
\newblock Multimodal learning and reasoning for visual question answering.
\newblock In {\em Advances in Neural Information Processing Systems},
  551--562.

\bibitem[\protect\citeauthoryear{Jiang \bgroup et al\mbox.\egroup
  }{2018}]{pythia2018}
Jiang, Y.; Natarajan, V.; Chen, X.; Rohrbach, M.; Batra, D.; and Parikh, D.
\newblock 2018.
\newblock Pythia v0.1: The winning entry to the vqa challenge 2018.
\newblock \url{https://github.com/facebookresearch/pythia}.

\bibitem[\protect\citeauthoryear{Kafle and Kanan}{2017}]{Kafle_2017_ICCV}
Kafle, K., and Kanan, C.
\newblock 2017.
\newblock An analysis of visual question answering algorithms.
\newblock In {\em The IEEE International Conference on Computer Vision (ICCV)}.

\bibitem[\protect\citeauthoryear{Kim \bgroup et al\mbox.\egroup
  }{2017}]{Kim2017}
Kim, J.-H.; On, K.~W.; Lim, W.; Kim, J.; Ha, J.-W.; and Zhang, B.-T.
\newblock 2017.
\newblock {Hadamard Product for Low-rank Bilinear Pooling}.
\newblock In {\em The 5th International Conference on Learning
  Representations}.

\bibitem[\protect\citeauthoryear{Kingma and Ba}{2015}]{KingmaB14}
Kingma, D.~P., and Ba, J.
\newblock 2015.
\newblock Adam: {A} method for stochastic optimization.
\newblock In {\em ICLR}.

\bibitem[\protect\citeauthoryear{Kiros \bgroup et al\mbox.\egroup
  }{2015}]{Kiros_2015_NIPS}
Kiros, R.; Zhu, Y.; Salakhutdinov, R.; Zemel, R.~S.; Torralba, A.; Urtasun, R.;
  and Fidler, S.
\newblock 2015.
\newblock Skip-thought vectors.
\newblock In {\em NIPS}.

\bibitem[\protect\citeauthoryear{Kiros, Salakhutdinov, and
  Zemel}{2015}]{kiros2015unifying}
Kiros, R.; Salakhutdinov, R.; and Zemel, R.~S.
\newblock 2015.
\newblock Unifying visual-semantic embeddings with multimodal neural language
  models.
\newblock {\em TACL}.

\bibitem[\protect\citeauthoryear{Krishna \bgroup et al\mbox.\egroup
  }{2017}]{Krishna_2017_IJCV}
Krishna, R.; Zhu, Y.; Groth, O.; Johnson, J.; Hata, K.; Kravitz, J.; Chen, S.;
  Kalantidis, Y.; Li, L.-J.; Shamma, D.~A.; et~al.
\newblock 2017.
\newblock Visual genome: Connecting language and vision using crowdsourced
  dense image annotations.
\newblock {\em International Journal of Computer Vision} 123(1):32--73.

\bibitem[\protect\citeauthoryear{Li \bgroup et al\mbox.\egroup
  }{2017}]{Li_2017_CVPR}
Li, Y.; Ouyang, W.; Wang, X.; and Tang, X.
\newblock 2017.
\newblock Vip-cnn: Visual phrase guided convolutional neural network.
\newblock In {\em CVPR}.

\bibitem[\protect\citeauthoryear{Liang, Lee, and Xing}{2017}]{Liang_2017_CVPR}
Liang, X.; Lee, L.; and Xing, E.~P.
\newblock 2017.
\newblock Deep variation-structured reinforcement learning for visual
  relationship and attribute detection.
\newblock In {\em CVPR}.

\bibitem[\protect\citeauthoryear{Lu \bgroup et al\mbox.\egroup
  }{2015}]{lu2015deeper}
Lu, J.; Lin, X.; Batra, D.; and Parikh, D.
\newblock 2015.
\newblock Deeper lstm and normalized cnn visual question answering model.

\bibitem[\protect\citeauthoryear{Lu \bgroup et al\mbox.\egroup
  }{2016}]{VRD_Lu_2016_ECCV}
Lu, C.; Krishna, R.; Bernstein, M.; and Fei-Fei, L.
\newblock 2016.
\newblock Visual relationship detection with language priors.
\newblock In {\em ECCV}.

\bibitem[\protect\citeauthoryear{Mordan \bgroup et al\mbox.\egroup
  }{2017}]{bmvc17}
Mordan, T.; Thome, N.; Henaff, G.; and Cord, M.
\newblock 2017.
\newblock Deformable part-based fully convolutional network for object
  detection.
\newblock In {\em BMVC}.

\bibitem[\protect\citeauthoryear{Noh and Han}{2016}]{Noh_2016_Arxiv}
Noh, H., and Han, B.
\newblock 2016.
\newblock Training recurrent answering units with joint loss minimization for
  vqa.
\newblock {\em arXiv preprint arXiv:1606.03647}.

\bibitem[\protect\citeauthoryear{Peyre \bgroup et al\mbox.\egroup
  }{2017}]{Peyre17}
Peyre, J.; Laptev, I.; Schmid, C.; and Sivic, J.
\newblock 2017.
\newblock Weakly-supervised learning of visual relations.
\newblock {\em ICCV}.

\bibitem[\protect\citeauthoryear{Teney \bgroup et al\mbox.\egroup
  }{2018}]{Teney_2018_CVPR}
Teney, D.; Anderson, P.; He, X.; and van~den Hengel, A.
\newblock 2018.
\newblock Tips and tricks for visual question answering: Learnings from the
  2017 challenge.
\newblock In {\em The IEEE Conference on Computer Vision and Pattern
  Recognition (CVPR)}.

\bibitem[\protect\citeauthoryear{Tucker}{1966}]{Tucker1966}
Tucker, L.~R.
\newblock 1966.
\newblock Some mathematical notes on three-mode factor analysis.
\newblock {\em Psychometrika} 31(3):279--311.

\bibitem[\protect\citeauthoryear{Yu \bgroup et al\mbox.\egroup
  }{2017a}]{Yu_2017_ICCV}
Yu, R.; Li, A.; Morariu, V.~I.; and Davis, L.~S.
\newblock 2017a.
\newblock Visual relationship detection with internal and external linguistic
  knowledge distillation.
\newblock In {\em ICCV}.

\bibitem[\protect\citeauthoryear{Yu \bgroup et al\mbox.\egroup
  }{2017b}]{yu2017mfb}
Yu, Z.; Yu, J.; Fan, J.; and Tao, D.
\newblock 2017b.
\newblock Multi-modal factorized bilinear pooling with co-attention learning
  for visual question answering.
\newblock {\em ICCV}.

\bibitem[\protect\citeauthoryear{Yu \bgroup et al\mbox.\egroup
  }{2018}]{yu2018beyond}
Yu, Z.; Yu, J.; Xiang, C.; Fan, J.; and Tao, D.
\newblock 2018.
\newblock Beyond bilinear: Generalized multi-modal factorized high-order
  pooling for visual question answering.
\newblock {\em IEEE TNNLS}.

\bibitem[\protect\citeauthoryear{Zhang, Hare, and
  Prügel-Bennett}{2018}]{zhang2018learning}
Zhang, Y.; Hare, J.; and Prügel-Bennett, A.
\newblock 2018.
\newblock Learning to count objects in natural images for visual question
  answering.
\newblock In {\em ICLR}.

\end{thebibliography}
\bibliographystyle{aaai}
\end{document}